\documentclass[letterpaper]{article} % DO NOT CHANGE THIS
\usepackage{aaai24}  % DO NOT CHANGE THIS
\usepackage{times}  % DO NOT CHANGE THIS
\usepackage{helvet}  % DO NOT CHANGE THIS
\usepackage{courier}  % DO NOT CHANGE THIS
\usepackage[hyphens]{url}  % DO NOT CHANGE THIS
\usepackage{graphicx} % DO NOT CHANGE THIS
\usepackage{natbib}  % DO NOT CHANGE THIS AND DO NOT ADD ANY OPTIONS TO IT
\usepackage{caption} % DO NOT CHANGE THIS AND DO NOT ADD ANY OPTIONS TO IT
\usepackage{algorithm}
\usepackage{algorithmic}

\usepackage{newfloat}
\usepackage{listings}
\DeclareCaptionStyle{ruled}{labelfont=normalfont,labelsep=colon,strut=off} % DO NOT CHANGE THIS
\floatstyle{ruled}
\newfloat{listing}{tb}{lst}{}
\floatname{listing}{Listing}
\usepackage{xcolor} 
\usepackage{algorithm}
\usepackage{algorithmic}
\usepackage{amsmath}
\usepackage{amssymb}
\usepackage{tabularx}
\usepackage{multirow}
\usepackage{rotating} % Required for rotating the table
\usepackage{caption}
\usepackage{subcaption}
\usepackage{pifont}
\usepackage{booktabs}

\usepackage{enumitem}
\usepackage{cleveref}
\usepackage{array}

\title{Progressive Text-to-Image Diffusion with Soft Latent Direction}

\author{
    Yuteng Ye,
    Jiale Cai,
    Hang Zhou, 
    Guanwen Li, 
    Youjia Zhang, 
    Zikai Song, 
    Chenxing Gao, 
    Junqing Yu,
    Wei Yang\thanks{indicates corresponding author.}
}
\affiliations{
    Huazhong University of Science and Technology, Wuhan, China\\
    \{yuteng\_ye, jaile\_cai, henrryzh, juggle\_lee, youjiazhang, skyesong, cxg, yjqing, weiyangcs\}@hust.edu.cn
}

\let\oldtwocolumn\twocolumn
\renewcommand\twocolumn[1][]{%
    \oldtwocolumn[{#1}{
    
\begin{center}
\vspace{-1.5em}
\includegraphics[width=0.98\textwidth]{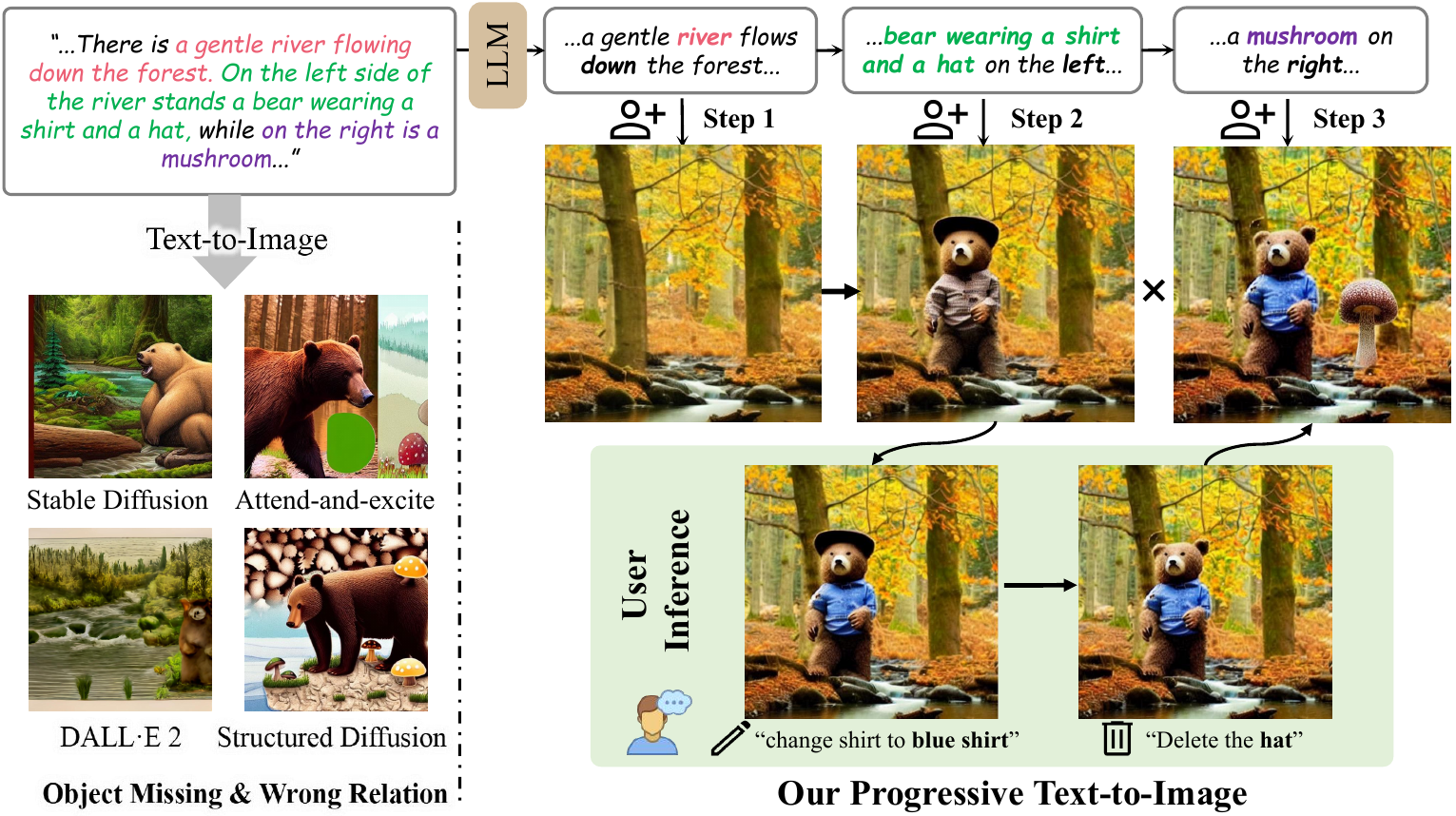}
    \captionof{figure}{
        Existing text-to-image synthesis approaches struggle with textual prompts involving multiple entities and specified relational directions. We propose to decompose the protracted prompt into a set of short commands, including synthesis, editing and erasing operations, using a Large Language Model (LLM) and progressively generate the image.
        Our strategy enhances both controllability and fidelity and allows for interactive modifications from user interference at each generation step.
    }
    \label{fig:teaser}
\end{center}
    }]
}

\begin{document}

\maketitle
%%%%%%%%% ABSTRACT
\begin{abstract}
In spite of the rapidly evolving landscape of text-to-image generation, the synthesis and manipulation of multiple entities while adhering to specific relational constraints pose enduring challenges. This paper introduces an innovative progressive synthesis and editing operation that systematically incorporates entities into the target image, ensuring their adherence to spatial and relational constraints at each sequential step.
Our key insight stems from the observation that while a pre-trained text-to-image diffusion model adeptly handles one or two entities, it often falters when dealing with a greater number. To address this limitation, we propose harnessing the capabilities of a Large Language Model (LLM) to decompose intricate and protracted text descriptions into coherent directives adhering to stringent formats.
To facilitate the execution of directives involving distinct semantic operations—namely insertion, editing, and erasing—we formulate the Stimulus, Response, and Fusion (SRF) framework. Within this framework, latent regions are gently stimulated in alignment with each operation, followed by the fusion of the responsive latent components to achieve cohesive entity manipulation.
Our proposed framework yields notable advancements in object synthesis, particularly when confronted with intricate and lengthy textual inputs. Consequently, it establishes a new benchmark for text-to-image generation tasks, further elevating the field's performance standards.

\end{abstract}

%%%%%%%%% BODY TEXT
\section{Introduction}
Text-to-image generation is a vital and rapidly evolving field in computer vision that has attracted unprecedented attention from both researchers and the general public.
The remarkable advances in this area are driven by the application of state-of-the-art image-generative models, such as auto-regressive~\cite{ramesh2021zero, wang2022clip} and diffusion models~\cite{ramesh2022hierarchical,saharia2022photorealistic,rombach2022high}, as well as the availability of large-scale language-image datasets~\cite{sharma2018conceptual, schuhmann2022laion}. However, existing methods face challenges in synthesizing or editing multiple subjects with specific relational and attributive constraints from textual prompts~\cite{chefer2023attend}.
The typical defects that occur in the synthesis results are missing entities, and inaccurate inter-object relations, as shown in ~\cref{fig:teaser}. Existing work improves the compositional skills of text-to-image synthesis models by incorporating linguistic structures~\cite{feng2022training}, and attention controls~\cite{hertz2022prompt,chefer2023attend} within the diffusion guidance process. 
Notably, Structured Diffusion~\cite{feng2022training} parse a text to extract numerous noun phrases, Attend-and-Excite~\cite{chefer2023attend} strength attention activations associated with the most marginalized subject token. Yet, these remedies still face difficulties when the text description is long and complex, especially when it involves two and more subjects. 
%Moreover, users may need to apply minor adjustment to unsatisfied regions of the generated image while preserving remaining parts.
Furthermore, users may find it necessary to perform subtle modifications to the unsatisfactory regions of the generated image, while preserving the remaining areas.

In this paper, we propose a novel progressive synthesizing/editing operation that successively incorporates entities, that conform to the spatial and relational constraint defined in the text prompt, while preserving the structure and aesthetics in each step. Our intuition is based on the observation that text-to-image models tend to better handle short-sentence prompts with a limited number of entities (1 or 2) than long descriptions with more entities. 
Therefore, we can parse the long descriptions into short-text prompts and craft the image progressively via a diffusion model to prevent the leakage and missing of semantics. 

However, applying such a progressive operation to diffusion models faces two major challenges:

\begin{itemize}[leftmargin=*]
\item The absence of a unified method for converting the integrated text-to-image process into a progressive procedure that can handle both synthesis and editing simultaneously. Current strategies can either synthesize~\cite{chefer2023attend, ma2023directed} or edit~\cite{kawar2023imagic,goel2023pair, xie2022smartbrush,avrahami2022blendedlatent, yang2023paint}, leaving a gap in the collective integration of these functions.

\item The need for precise positioning and relational entity placement. Existing solutions either rely on user-supplied masks for entity insertion, necessitating manual intervention~\cite{avrahami2022blendedlatent, nichol2021glide}, or introduce supplementary phrases to determine the entity editing direction~\cite{hertz2022prompt,brooks2023instructpix2pix}, which inadequately addressing spatial and relational dynamics.

\end{itemize}

To overcome these hurdles, we present the Stimulus, Response, and Fusion (SRF) framework, assimilating a stimulus-response generation mechanism along with a latent fusion module into the diffusion process. Our methodology involves employing a fine-tuned GPT model to deconstruct complex texts into structured prompts, including synthesis, editing, and erasing operations governed by a unified SRF framework.
Our progressive process begins with a real image or synthesized background, accompanied by the text prompt, and applies the SRF method in a step-by-step approach. Unlike previous strategies that aggressively manipulate the cross-attention map~\cite{wu2023harnessing,ma2023directed}, our operation guides the attention map via a soft direction, avoiding brusque modifications that may lead to discordant synthesis.
Additionally, when addressing relationships like ``wearing'' and ``playing with'', we begin by parsing the positions of the objects, after which we incorporate the relational description into the diffusion process to enable object interactions.

In summary, we unveil a novel, progressive text-to-image diffusion framework that leverages the capabilities of a Language Model (LLM) to simplify language description, offering a unified solution for handling synthesis and editing patterns concurrently. This represents an advancement in text-to-image generation and provides a new platform for future research.

 % --- layout-pipeline ---
\begin{figure}[t]
    \centering
    \includegraphics[width=0.45\textwidth]{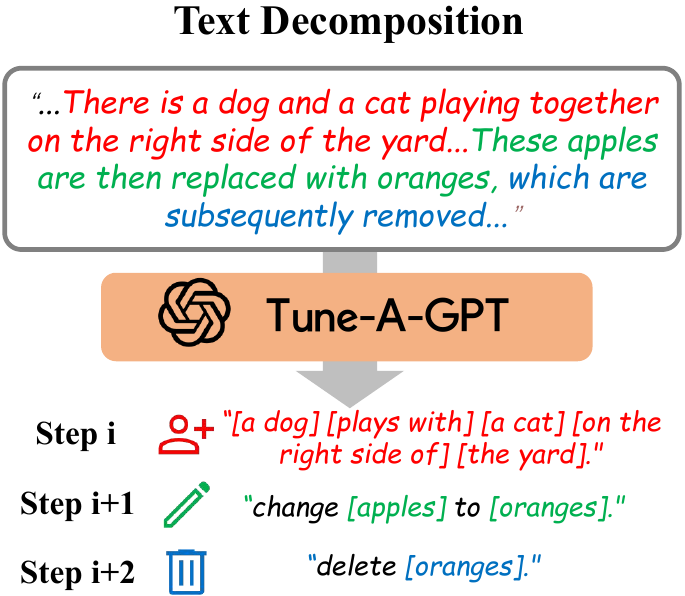}
    \caption{We employ a fine-tuned GPT model to deconstruct a comprehensive text into structured prompts, each classified under synthesis, editing, and erasing operations.}
    \label{fig:layout-pipeline}
\end{figure}

\begin{figure}[t]
    \centering
    \includegraphics[width=0.45\textwidth]{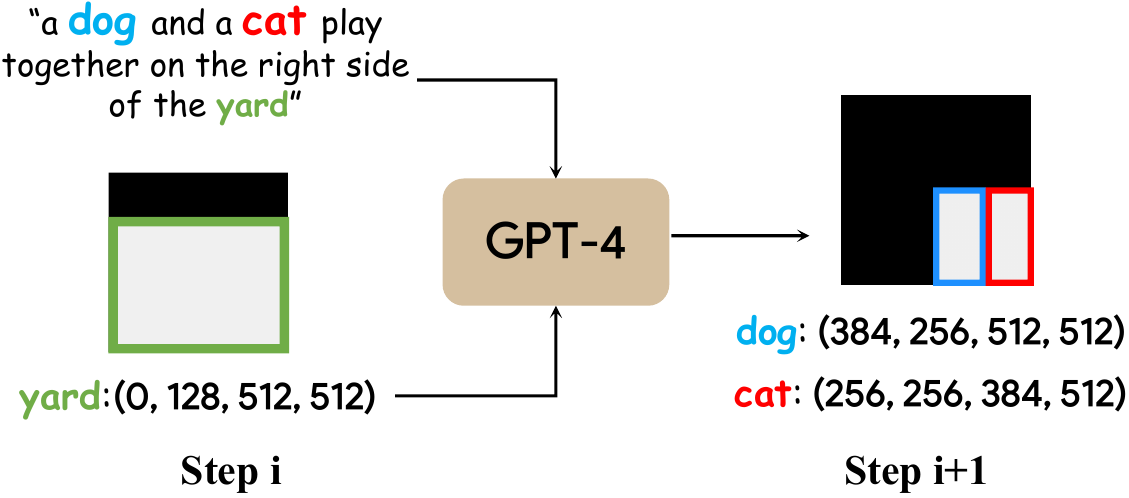}
    \caption{For the synthesis operation, we generate the layout indicated in the prompt from a frozen GPT-4 model, which subsequently yields the new bounding box coordinates for object insertion.}
    \label{fig:synthesis-layout-pipeline}
\end{figure}

%  --- pipleline ---
\begin{figure*}[t]
    \centering
    \includegraphics[width=0.95\textwidth]{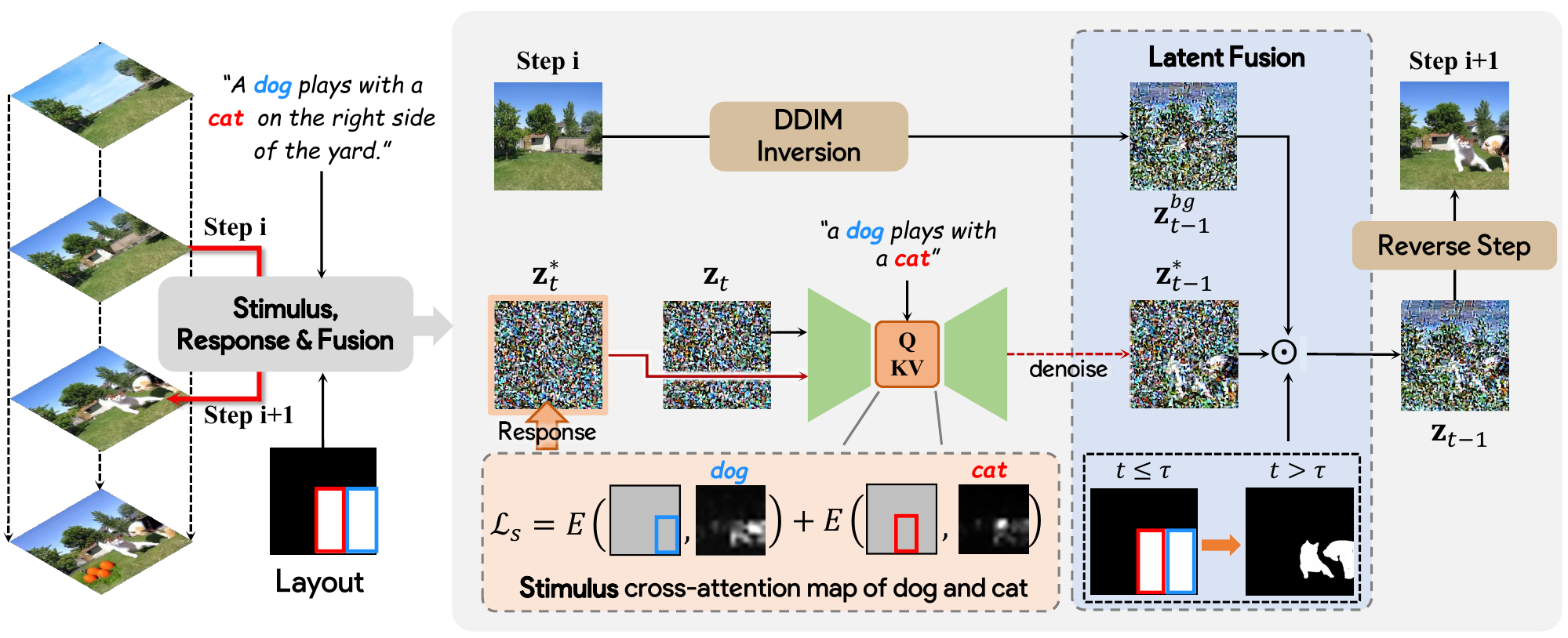}
    \caption{Overview of our unified framework emphasizes progressive synthesis, editing, and erasing. In each progressive step, A random latent \(z_t\) is directed through the cross-attention map in inverse diffusion. Specifically, we design a soft stimulus loss that evaluates the positional difference between entity attention and the target mask region, leading to a gradient for updating the latent \(z_{t-1}^{*}\) as a latent response. Subsequentially, another forward diffusion pass is applied to denoise \(z^*_{t}\), yielding deriving \(z^{*}_{t-1}\). In the latent fusion phase, we transform the previous \(i\)-th image into a latent code \(z^{bg}_{t-1}\) using DDIM inversion. The blending of \(z^{*}_{t-1}\) with \(z^{bg}_{t-1}\) incorporates a dynamic evolving mask, which starts with a layout box and gradually shifts to cross-attention. Finally, \(z^{*}_{t-1}\) undergoes multiple diffusion reverse steps and results in the \((i+1)\)-th image.}
    \label{fig:pipeline}
\end{figure*}

%-------------------------------------------------------------------------
\section{Related Work}
% Our method is closely related to image manipulation and cross-attention control within diffusion models.

\subsection{Image Manipulation}
Image manipulating refers to the process of digitally manipulating images to modify or enhance their visual appearance. Various techniques can be employed to achieve this end, such as the use of spatial masks or natural language descriptions to guide the editing process towards specific goals. One promising line of inquiry involves the application of generative adversarial networks (GANs) for image domain transfer~\cite{isola2017image,sangkloy2017scribbler,zhu2017unpaired,choi2018stargan,wang2018high,huang2018multimodal,park2019semantic,liu2017unsupervised,baek2021rethinking} or the manipulation of latent space~\cite{zhu2016generative,huh2020transforming,richardson2021encoding,zhu2020domain,wulff2020improving,bau2021paint}. 
Recently, diffusion models have emerged as the mainstream. GLIDE~\cite{nichol2021glide}, Blended diffusion~\cite{avrahami2022blendedlatent} and SmartBrush~\cite{xie2022smartbrush} replace masked image regions with predefined objects while preserving the inherent image structure. Additionally, techniques such as prompt-to-prompt~\cite{hertz2022prompt} and instructpix2pix~\cite{brooks2023instructpix2pix} enable the modification of image-level objects through text alterations.
Contrasting previous methods that solely cater to either synthesis or editing, we construct a unified framework that accommodates both.

\subsection{Cross Attention Control}
Objects and positional relationships are manifested within the cross attention map of the diffusion model. Inspired by this observation~\cite{feng2022training}, techniques have been devised to manipulate the cross attention map for image synthesis or editing.
Prompt-to-Prompt approach~\cite{hertz2022prompt} aims at regulating spatial arrangement and geometry through the manipulation of attention maps derived from textual prompts.
Structured Diffusion~\cite{feng2022training} utilizes a text parsing mechanism to isolate numerous noun phrases, enhancing the corresponding attention space channels.
The Attend-and-Excite approach~\cite{chefer2023attend} amplifies attention activations linked to the most marginalized subject tokens.
Directed Diffusion~\cite{ma2023directed} proposes an attention refinement strategy through the utilization of a weak and strong activation approach.
% 
% Our approach stands apart by guiding the attention through Soft Latent Direction. Instead of just amplifying or refining attention activations, our soft latent direction serves as a gentle guide, ensuring smoother and more natural transitions in the attention space.
% 
The main difference between our layout generation and the layout prediction approaches is that our method enables precise increment generation and intermediate modifications, i.e., we gradually change the layout instead of generating one layout at once. As for background fusion, we use a soft mask to ensure the object's integrity. 

% In contrast, we harness attention responses to actualize positional and relational phases. Additionally, we employ latent fusion for the seamless integration of elements into pre-existing images.
\section{Method}
\subsection{Problem Formulation}
we elaborate upon our innovative progressive text-to-image framework. Given a multifaceted text description \(\mathcal{P}\) and a real or generated background \(\mathcal{I}\), our primary goal is to synthesize an image that meticulously adheres to the modifications delineated by \(\mathcal{P}\) in alignment with  \(\mathcal{I}\).
The principal challenge emerges from the necessity to decode the intricacy of \(\mathcal{P}\), manifesting across three complex dimensions:
\begin{itemize}
\item The presence of multiple entities and attributes escalates the complexity of the scene, imposing stringent demands on the model to generate representations that are not only accurate but also internally coherent and contextually aligned.
\item The integration of diverse positional and relational descriptions calls for the model to exhibit an advanced level of understanding and to employ sophisticated techniques to ascertain precise spatial configuration, reflecting both explicit commands and implied semantic relations.
\item The concurrent introduction of synthesis, editing, and erasing operations introduces additional layers of complexity to the task. Managing these intricate operations within a unified model presents a formidable challenge, requiring a robust and carefully designed approach to ensure seamless integration and execution.
\end{itemize}

We address these challenges through a unified progressive text-to-image framework that: (1) employs a fine-tuned GPT model to distill complex texts into short prompts, categorizing each as synthesis, editing, or erasing mode, and accordingly generating the object mask; (2) sequentially processes these prompts within the same framework, utilizing attention-guided generation to capture position-aware features with soft latent direction, and subsequently integrates them with the previous stage's outcomes in a subtle manner. This approach synthesizes the intricacies of text-to-image transformation into a coherent, positionally aware procedure.

\subsection{Text Decomposition}
\(\mathcal{P}\) may involve multiple objects and relations, we decompose \(\mathcal{P}\) into a set of short prompts, which produces an image accurately representing \(\mathcal{P}\) when executed sequentially.
As illustrated in \cref{fig:layout-pipeline}, we fine-tune a GPT with OpenAI API~\cite{openai2023gpt4} to decompose \(\mathcal{P}\) into multiple structured prompts, denoted as $\{\mathcal{P}_1, \mathcal{P}_2, ..., \mathcal{P}_n \}$.
Each \(\mathcal{P}_i\) falls into one of the three distinct modes: 
\textbf{Synthesis mode:} ``[\textcolor{gray}{object 1}] [\textcolor{gray}{relation}] [\textcolor{gray}{object 2}] [\textcolor{gray}{position}] [\textcolor{gray}{object 3}]'', 
\textbf{Editing mode:} ``change [\textcolor{gray}{object 1}] to [\textcolor{gray}{object 2}]'', 
and \textbf{Erasing mode:} ``delete [\textcolor{gray}{object}]''.
In pursuit of this aim, we start by collecting full texts using ChatGPT~\cite{brown2020language} and then manually deconstruct them into atomic prompts. Each prompt has a minimal number of relations and is labeled with synthesis/editing/erasing mode. Using these prompts and their corresponding modes for model supervision, we fine-tune the GPT model to enhance its decomposition and generalization ability.

% Layout Synthesis Module
% Attention Acquisition Module

\textbf{Operational Layouts.}
% As illustrated in \cref{fig:layout-pipeline}, we introduce the \textbf{Layout Synthesis Module} and \textbf{Attention Acquisition Module} for the synthesis and editing operations, respectively.
% 
For the synthesis operation, as shown in \cref{fig:synthesis-layout-pipeline}, we feed both the prompt and a reference bounding box into a frozen GPT-4 API. This procedure produces bounding boxes for the target entity that will be used in the subsequent phase. We exploit GPT-4's ability to extract information from positional and relational text descriptors. For example, the phrase ``cat and dog play together'' indicates a close spatial relationship between the ``cat'' and ``dog''. Meanwhile, ``on the right side'' suggests that both animals are positioned to the right of the ``yard''.
For the editing and erasing operations, we employ Diffusion Inversion~\cite{mokady2023null} to obtain the cross-attention map of the target object, which serves as the layout mask. For example, when changing ``apples'' to ``oranges'', we draw upon the attention corresponding to ``apples''. On the other hand, to ``delete the oranges'', we focus on the attention related to ``oranges''. Notably, this approach avoids the need to retrain the diffusion model and is proficient in managing open vocabularies.
we denote generated layout mask as \(\mathcal{M}\) for all operations in following sections for convention.

In the following section, we provide a complete introduction to the synthesis operation. At last, we exhibit that the editing and erasing operations only differ from the synthesis operation in parameter settings.
% This framework contains stimulus \& response as well as latent fusion.

\subsection{Stimulus \& Response}
\begin{figure}[h]
    \centering
    \includegraphics[width=0.45\textwidth]{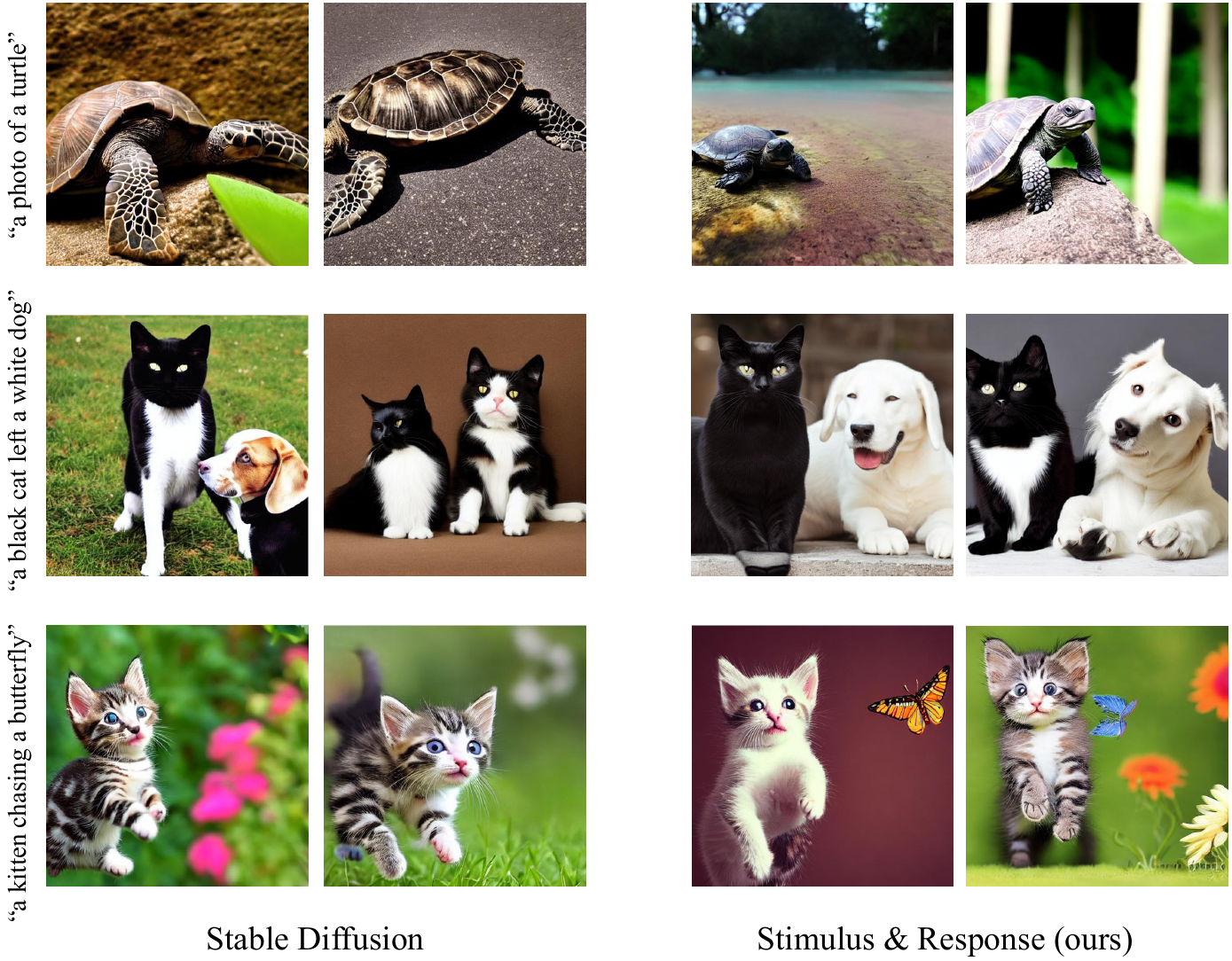}
    \caption{Visual results generated by Stable Diffusion and Stimulus \& Response. Stable Diffusion shows noticeable problems in positional generation (top), semantic and attribute coupling (middle), and object omission (bottom), while ours delivers precise outcomes.}
    \label{fig:Stimulus_and_Response_with_SD}
\end{figure}

With the synthesis prompt $\mathcal{P}_i$ to be executed and its mask configuration $\mathcal{M}_i$. The goal of Latent Stimulus \& Response is to enhance the positional feature representation on \(\mathcal{M}\). As illustrated in \cref{fig:pipeline}, this is achieved by guided cross-attention generation. 
Differing from the approaches~\cite{ma2023directed, wu2023harnessing}, which manipulate attention through numerical replacement, we modulate the attention within mask regions associated with the entity in $\mathcal{P}_i$ via a soft manner. Rather than directly altering the attention, we introduce a stimulus to ensure that the object attention converges to the desired scores.
Specifically, we formulate a stimulus loss function between the object mask \(\mathcal{M}\) and the corresponding attention \(A\) as:
\begin{equation}
\mathcal{L}_{s} = \sum_{i=1}^{n} (\text{softmax}(A^i_{t}) - \delta \cdot \mathcal{M}^i)
\label{eq:stimulus loss}
\end{equation}
\noindent where $A^i_t$ signifies the cross-attention map of the $i$-th object at the $t$-th timestep. $\mathcal{M}^i$ denotes the mask of the $i$-th object. $\delta$ represents the stimulus weights.
The intent of stimulus attention leans towards a spatial-wise generation process. This is achieved by backpropagating the gradient of the stimulus loss function, as defined in Eq.~\ref{eq:stimulus loss}, to update the latent code. This process serves as a latent response to the stimulated attention, which can be formally expressed as:
\begin{equation}
z^{*}_{t} \gets z_t - \alpha_t \cdot \nabla_{z_t} \mathcal{L}_{s}
\label{eq-alpha}
\end{equation}
\noindent In the above equation, $z^{*}_{t}$ represents the updated latent code and $\alpha_t$ denotes the learning rate. Finally, we execute another forward pass of the stable diffusion model using the updated latent code $z^*_{t}$ to compute $z^{*}_{t-1}$ for the subsequent denoising step.
Based on \cref{eq:stimulus loss} and \cref{eq-alpha}, we observe consistent spatial behavior in both the cross-attention and latent spaces. For a more detailed analysis, we refer to \cref{fig:Stimulus_and_Response_with_SD} and find this property contributes to producing faithful and position-aware image representations.

% Moreover, a comparative analysis between the Stimulus & Response mechanism and Stable Diffusion, illustrated in \cref{fig:Stimulus_and_Response_with_SD}, reveals that this consistency addresses challenges like incorrect positional generation, semantic and attribute interplay, and object omission commonly observed in Stable Diffusion.

%  --- Stimulus \& Response results ---

\subsection{Latent Fusion}
Recalling that \(z^*_{t-1}\) denotes the latent feature of the target object, our next task is to integrate them seamlessly with the image from the preceding stage. For this purpose, we first convert the previous image into latent code by DDIM inversion, denoted as \(z^{bg}\). Then for timestep t, we take a latent fusion strategy~\cite{avrahami2022blended} between \(z^{bg}_t\) and \(z^{*}_t\), which is formulated as:
\begin{equation}
z_{t-1} = \mathcal{\widehat{M}} \cdot z^*_{t-1} + (1 - \mathcal{\widehat{M}}) \cdot z^{bg}_{t-1}
\label{latent mask}
\end{equation}
\noindent where \(\mathcal{\widehat{M}}\) acts as a latent mask to blend the features of target objects with the background. In the synthesis operation, employing a uniform mask across all steps can be too restrictive, potentially destroying the object's semantic continuity. To mitigate this, we introduce a more soft mask, ensuring both object integrity and spatial consistency. Specifically, during the initial steps of diffusion denoising, we use layout mask \(\mathcal{M}\) to provide spatial guidance. Later, we shift to an attention mask \(\mathcal{M}_{\text{attn}}\), generated by averaging and setting a threshold on the cross-attention map, to maintain object cohesion. This process is denoted as:

\begin{align} 
\mathcal{\widehat{M}}(\mathcal{M}_{\text{attn}}, \mathcal{M}, t) = 
\begin{cases}
\mathcal{M} & \text{if } t \leq \tau \\
\mathcal{M}_{\text{attn}} & \text{if } t > \tau
\end{cases}
\label{eq:latent-fusion-number}
\end{align}

\noindent Here, \(\tau\) serves as a tuning parameter balancing object integrity with spatial coherence. 
The above response and fusion process is repeated for a subset of the diffusion timesteps, and the final output serves as the image for the next round generation.

\textbf{Editing and Erasing Specifications.} Our editing and erasing operation differs in parameter setting: we set \(\mathcal{M}\) in \cref{eq:stimulus loss} as editing/erasing reference attention. we set \(\mathcal{\widehat{M}}\) in \cref{latent mask} as the editing/erasing mask in all diffusion steps for detailed, shape-specific modifications.

\section{Experiment}

\textbf{Baselines and Evaluation.}
Our experimental comparison primarily concentrates on Single-Stage Generation and Progressive Generation baselines.
(1) We refer to \textbf{Single-Stage Generation} methods as those that directly generate images from input text in a single step. Current methods include Stable Diffusion~\cite{rombach2022high}, Attend-and-excite~\cite{chefer2023attend}, and Structured Diffusion~\cite{feng2022training}. We compare these methods to analyze the efficacy of our progressive synthesis operation. We employ GPT to construct 500 text prompts that contain diverse objects and relationship types.
For evaluation, we follow~\cite{wu2023harnessing} to compute \textbf{Object Recall}, which quantifies the percentage of objects successfully synthesized. Moreover, we measure \textbf{Relation Accuracy} as the percentage of spatial or relational text descriptions that are correctly identified, based on 8 human evaluations.
(2) We define \textbf{Progressive Generation} as a multi-turn synthesis and editing process that builds on images from preceding rounds. Our comparison encompasses our comprehensive progressive framework against other progressive methods, which includes Instruct-based Diffusion models~\cite{brooks2023instructpix2pix} and mask-based diffusion models~\cite{rombach2022high,avrahami2022blendedlatent}. 
To maintain a balanced comparison, we source the same input images from SUN~\cite{xiao2016sun} and text descriptions via the GPT API~\cite{openai2023gpt4}. Specifically, we collate five scenarios totaling 25 images from SUN, a dataset that showcases real-world landscapes. Each image is paired with the text description, which ensures: 1. Integration of synthesis, editing, and easing paradigms; 2. Incorporation of a diverse assortment of synthesized objects; 3. Representation of spatial relations (e.g., top, bottom, left, right) and interactional relations (e.g., ``playing with'', ``wearing'').
For evaluation, we utilize Amazon Mechanical Turk (AMT) to assess image fidelity. Each image is evaluated based on the fidelity of the generated objects, their relationships, the execution of editing instructions, and the alignment of erasures with the text descriptions.
Images are rated on a fidelity scale from 0 to 2, where 0 represents the lowest quality and 2 signifies the highest. With two evaluators assessing each generated image, the cumulative score for each aspect can reach a maximum of 100.

%------------------------- result_one_stage ---------------------------
\begin{figure*}[h]
    \centering
    \includegraphics[width=1.0\textwidth]{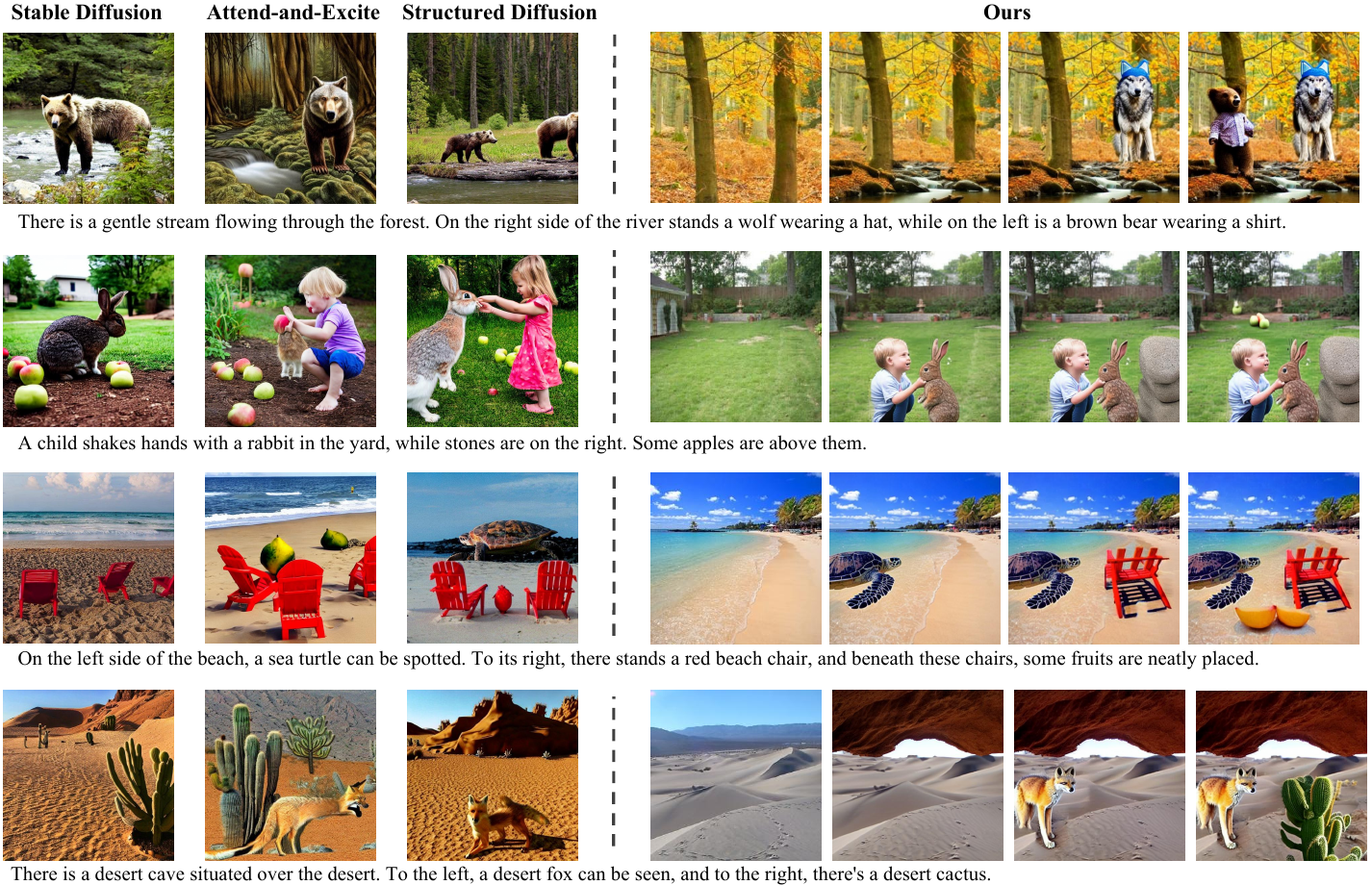}
    \caption{Qualitative comparison with Single-Stage baselines. Common errors in the baselines include missing objects and mismatched relations. Our method demonstrates the progressive generation process.}
    \label{fig:one_stage_comparison}
\end{figure*}

%------------------------- ablation_editing_for_attn_loss ---------------------------
\begin{figure*}[t]
    \centering
    \includegraphics[width=0.95\textwidth]{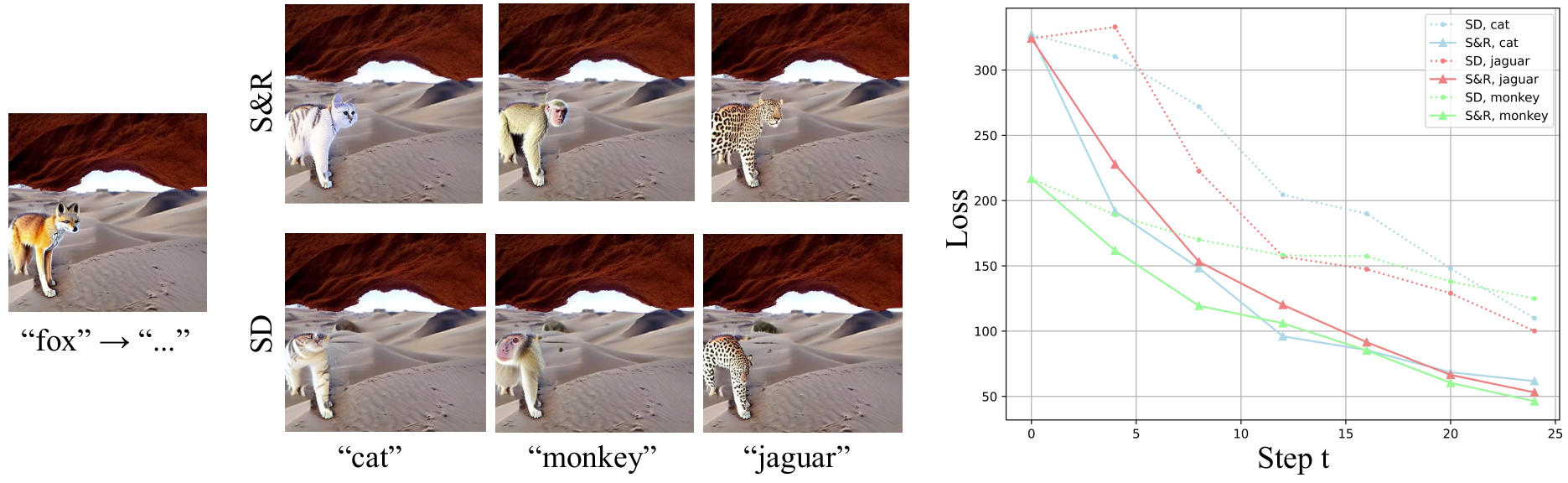} 
    \caption{The analysis of Stimulus \& Response in the editing operation. The left side shows a visual comparison between SD (Stable Diffusion) and S\&R (Stimulus \& Response). The right side presents the convergence curve of cross-attention loss during diffusion sampling steps. The loss is computed as the difference between reference attention and model-generated attention. In the right figure, red, blue, and green colors represent the objects ``jaguar'', ``cat'', and ``monkey'' respectively. Solid lines indicate SD loss, while dashed lines represent S\&D loss.}
    \label{fig:Ablation_SR_editing}
\end{figure*}

%------------------------- prog_result ---------------------------
\begin{figure}[p]
    \centering
    \includegraphics[width=0.45\textwidth]{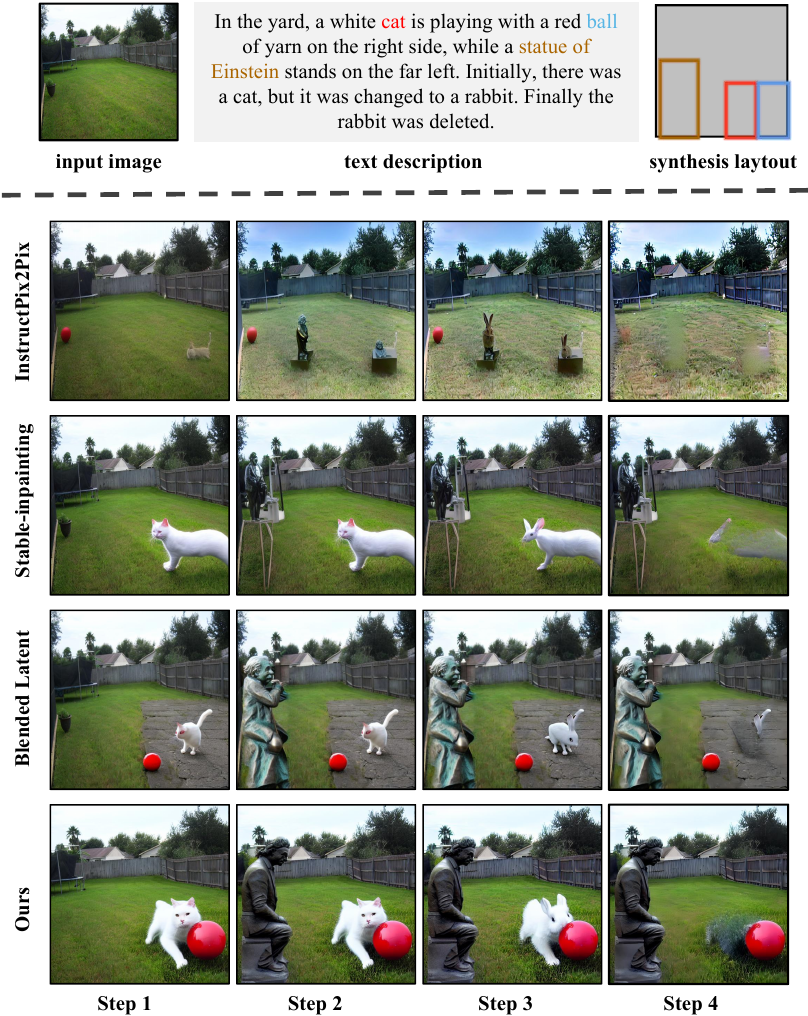} 
    \caption{Qualitative comparison with Progressive Generation baselines. The first two phases illustrate object synthesis operation, where target objects are color-coded in both the text and layout. Subsequent phases depict object editing and erasing processes, wherein a cat is first transformed into a rabbit and then the rabbit is removed.}
    \label{fig:progressive-generation}
\end{figure}

\textbf{Implementation Details.} 
Our framework builds upon Stable Diffusion (SD) V-1.4. During the Stimulus \& Response stage, we assign a weight of \(\delta\) equals 0.8 in \cref{eq:stimulus loss}, and set \(t\) equals 25 and \(\alpha_t\) equals 40 in~\cref{eq-alpha}. We implement the stimulus procedure over the 16 × 16 attention units and integrate the Iterative Latent Refinement design~\cite{chefer2023attend}. In the latent fusion stage, the parameter \(\tau\) is set to a value of 40.

\subsection{Qualitative and Quantitative Results}

\textbf{Qualitative and Quantitative Comparisons with Single-Generation Baselines. } \cref{fig:one_stage_comparison} reveals that traditional baseline methods often struggle with object omissions and maintaining spatial and interactional relations. In contrast, our progressive generation process offers enhanced image fidelity and controllability. Additionally, we maintain finer details in the generated images, such as the shadows of the ``beach chair''. Result in \cref{tab:one_stage_comparison} indicates that our method outperforms the baselines in both object recall and relation accuracy.

\textbf{Qualitative and Quantitative Comparisons with Progressive Generation Baselines.} In \cref{fig:progressive-generation}, baseline methods often fail to synthesize full objects and may not represent relationships as described in the provided text. Moreover, during editing and erasing operations, these methods tend to produce outputs with compromised quality, showcasing unnatural characteristics. It's worth noting that any missteps or inaccuracies in the initial stages, such as those seen in InstructPix2Pix, can cascade into subsequent stages, exacerbating the degradation of results. In contrast, our proposed method consistently yields superior results through every phase. The results in \cref{tab:multi-turn-result} further cement our method's dominant performance in synthesis, editing, and erasing operations, as underscored by the impressive rating scores.

\begin{table}[t]
\small
    \centering
    \begin{tabular}{ccc}
        \hline
        \textbf{Method} & Object Recall~$\uparrow$ & Relation Accuracy~$\uparrow$ \\
        \hline
        Stable Diffusion & 40.7 & 19.8 \\
        \hline
        Structured Diffusion & 43.5 & 21.6 \\
        \hline
        Attend-and-excite & 50.3 & 23.4 \\
        \hline
        Ours & \textbf{64.4} & \textbf{50.8} \\
        \hline
    \end{tabular}
    \caption{Quantitative comparison with Single-Stage Generation baselines.}
    \label{tab:one_stage_comparison}
\end{table}

\begin{table}[t]
\small
    \centering
    % \resizebox{\linewidth}{!}{%
    \begin{tabular}{lccccc}
        \hline
        \multicolumn{1}{c}{\textbf{Method}} & \multicolumn{2}{c}{\textbf{Synthesis}} & \multicolumn{1}{c}{\textbf{Editing}} & \multicolumn{1}{c}{\textbf{Erasing}} \\
        \cmidrule(r){2-3}
        & Object & Relation & & \\
        \hline
        InstructPix2Pix & 19 & 24 & 32 & 29 \\
        \hline
        Stable-inpainting & 64 & 54 & 65 & 45 \\
        \hline
        Blended Latent & 67 & 52 & 67 & 46 \\
        \hline
        Ours & \textbf{74} & \textbf{60} & \textbf{72} & \textbf{50} \\
        \hline
    \end{tabular}%
    % }
    \caption{Quantitative comparison of our method against Progressive Generation baselines, using rating scores.}
    \label{tab:multi-turn-result}
\end{table}

\begin{table}[t]
\small
    \centering
    % \resizebox{0.8\linewidth}{!}{%
    \begin{tabular}{ccc}
        \hline
        \textbf{Method Variant} & Object Recall~$\uparrow$ & Relation Accuracy~$\uparrow$ \\
        \hline
        w/o LF & 38.8 & 21.8 \\
        \hline
        w/o S\&R & 58.3 & 45.2 \\
        \hline
        Ours & \textbf{64.4} & \textbf{50.8} \\
        \hline
    \end{tabular}%
    % }
    \caption{Ablation study. LF and S\&R represent Latent Fusion and Stimulus \& Response respectively.}
    \label{tab:ablation_study}
\end{table}

\subsection{Ablation Study}
\textbf{Ablation study of method components is shown in \cref{tab:ablation_study}.} Without latent fusion, we lose continuity from prior generation stages, leading to inconsistencies in object synthesis and placement. On the other hand, omitting the Stimulus \& Response process results in a lack of positional awareness, making the synthesis less precise. Both omissions manifest as significant drops in relation and entity accuracies, emphasizing the synergistic importance of these components in our approach.

\textbf{The analysis of Stimulus \& Response in the editing operation is highlighted in \cref{fig:Ablation_SR_editing}}. Compared to Stable Diffusion, Stimulus \& Response not only enhances object completeness and fidelity but also demonstrates a broader diversity in editing capabilities. The loss curve indicates that Stimulus \& Response aligns more closely with the reference cross-attention, emphasizing its adeptness in preserving the original structure.

\section{Conclusion}
In this study, we addressed the prevailing challenges in the rapidly advancing field of text-to-image generation, particularly the synthesis and manipulation of multiple entities under specific constraints. Our innovative progressive synthesis and editing methodology ensures precise spatial and relational representations. Recognizing the limitations of existing diffusion models with increasing entities, we integrated the capabilities of a Large Language Model (LLM) to dissect complex text into structured directives. Our Stimulus, Response, and Fusion (SRF) framework, which enables seamless entity manipulation, represents a major stride in object synthesis from intricate text inputs. 
%his work not only underscores the potential of our approach but also sets a new standard in text-to-image generation tasks

% {\color{red} One major limitation of our approach is that not all text can be decomposed into a sequence of short prompts. For instance, given text as ``a horse under a car and between a cat and a dog''. In this case, our approach will fail to parse the synthesis object and relation in gradually sequence. We plan to use more texts to acquire GPT parse ability to more diverse text prompts.}

One major limitation of our approach is that not all text can be decomposed into a sequence of short prompts. For instance, our approach finds it challenging to sequentially parse text such as ``a horse under a car and between a cat and a dog.'' We plan to gather more training data and labels of this nature to improve the parsing capabilities of GPT.

% \section{Acknowledgment}
% \textcolor{red}{
%     This work is supported by the National Key Research and Development Program of China under Grant No.2020YFB1805601, National Natural Science Foundation of China (NSFC No. 62272184), and CCF-Tencent Open Research Fund (CCF-Tencent RAGR20220120). The computation is completed in the HPC Platform of Huazhong University of Science and Technology.
% }
\section{Acknowledgments}
This work is supported by the National Natural Science Foundation of China (NSFC No. 62272184). The computation is completed in the HPC Platform of Huazhong University of Science and Technology.

\bibliography{aaai24}

\end{document}